\documentclass[letterpaper, 10 pt, journal, twoside]{IEEEtran}

\usepackage{url}

\usepackage{xspace}
\newcommand{\argmax}{\operatornamewithlimits{argmax}}

\newcommand{\algoName}{PlaTe\xspace} 

\providecommand{\jk}[1]{{\color{black} #1}}

\usepackage{graphics} 
\usepackage{epsfig} 
\usepackage{mathptmx} 
\usepackage{times} 
\usepackage{amsmath} 
\usepackage{amssymb}  

\usepackage[dvipsnames]{xcolor}
\usepackage{algorithm}
\usepackage{algorithmicx}
\usepackage{algpseudocode}
\usepackage{amsmath}
\usepackage{bm}

\usepackage{booktabs}
\usepackage{subcaption}
\usepackage{hyperref}
\definecolor{brightturquoise}{rgb}{0.03, 0.91, 0.87}
\usepackage{multirow}
\usepackage{array}

\begin{document}
\title{
\algoName: Visually-Grounded Planning 
with Transformers in Procedural Tasks}

\author{Jiankai Sun$^{1}$, De-An Huang$^{2}$, Bo Lu$^{3}$, Yun-Hui Liu$^{4}$, Bolei Zhou$^{5}$, Animesh Garg$^{2,6}$%
\thanks{Manuscript received: September 9, 2021; Revised December 7, 2021; Accepted January 10, 2022.}
\thanks{This paper was recommended for publication by Editor Cesar Cadena upon evaluation of the Associate Editor and Reviewers' comments.}
\thanks{$^{1} $Stanford University, Stanford, California, United States of America, 94305.}%
\thanks{$^{2} $NVIDIA, Cupertino, CA, United States of America, 95014.}%
\thanks{$^{3} $Robotics and Microsystems Center, School of Mechanical and Electric Engineering, Soochow University, Suzhou, Jiangsu, China.} 
\thanks{$^{4} $The Chinese University of Hong Kong, Hong Kong.}%
\thanks{$^{5} $University of California, Los Angeles, USA.}%
\thanks{$^{6} $University of Toronto \& Vector Institute, Canada.}%
}

\markboth{IEEE Robotics and Automation Letters. Preprint Version. Accepted January, 2022}
{Sun \MakeLowercase{\textit{et al.}}: \algoName: Visually-Grounded Planning 
with Transformers in Procedural Tasks} 

\maketitle
\begin{abstract}
In this work, we study the problem of how to leverage instructional videos to facilitate the understanding of human decision-making processes, focusing on training a model with the ability to plan a goal-directed procedure from real-world videos. Learning structured and plannable state and action spaces directly from unstructured videos is the key technical challenge of our task. There are two problems: first, the appearance gap between the training and validation datasets could be large for unstructured videos; second, these gaps lead to decision errors that compound over the steps.  We address these limitations with  \underline{Pla}nning \underline{T}ransform\underline{e}r (\algoName), which has the advantage of circumventing the compounding prediction errors that occur with single-step models during long model-based rollouts. Our method simultaneously learns the latent state and action information of assigned tasks and the representations of the decision-making process from human demonstrations.
Experiments conducted on real-world instructional videos show that our method can achieve a better performance in reaching the indicated goal than previous algorithms. We also validated the possibility of applying procedural tasks on a UR-5 platform. Please see \href{http://pair.toronto.edu/plate-planner}{pair.toronto.edu/plate-planner} for additional details.
\end{abstract}


\IEEEpeerreviewmaketitle

\section{Introduction}
\IEEEPARstart{I}{ntelligent} reasoning in embodied environments requires that an agent has explicit representations of parts or aspects of its environment to reason about~\cite{beetz2016ai}. 
As a generic reasoning application, action planning and learning are crucial skills for cognitive robotics.
Planning, in the traditional AI sense, means deliberating about a course of actions for an agent to take for achieving a given set of goals. The desired plan is a set of actions whose execution transforms the initial situation into the goal situation (goal situations need not be unique). Normally, actions in a plan cannot be executed in arbitrary sequence, but have to obey an ordering, ensuring that all preconditions of each action are valid at the time of its execution.
In practice, there are two challenges for planning. First, typically not all information that would be needed is available. Planning is meant for real environments in which many parameters are unknown or unknowable. Second, even if everything for a complete planning were known, then it would very likely be so computationally intensive that it would run too slowly in the real world. Thus, many planning systems and their underlying planning algorithms accept the restrictive assumptions of information completeness, determinism, instantaneousness, and idleness~\cite{beetz2016ai,pmlr-v155-sun21a}.

\begin{figure}[t]
\centering
  \includegraphics[width=\linewidth]{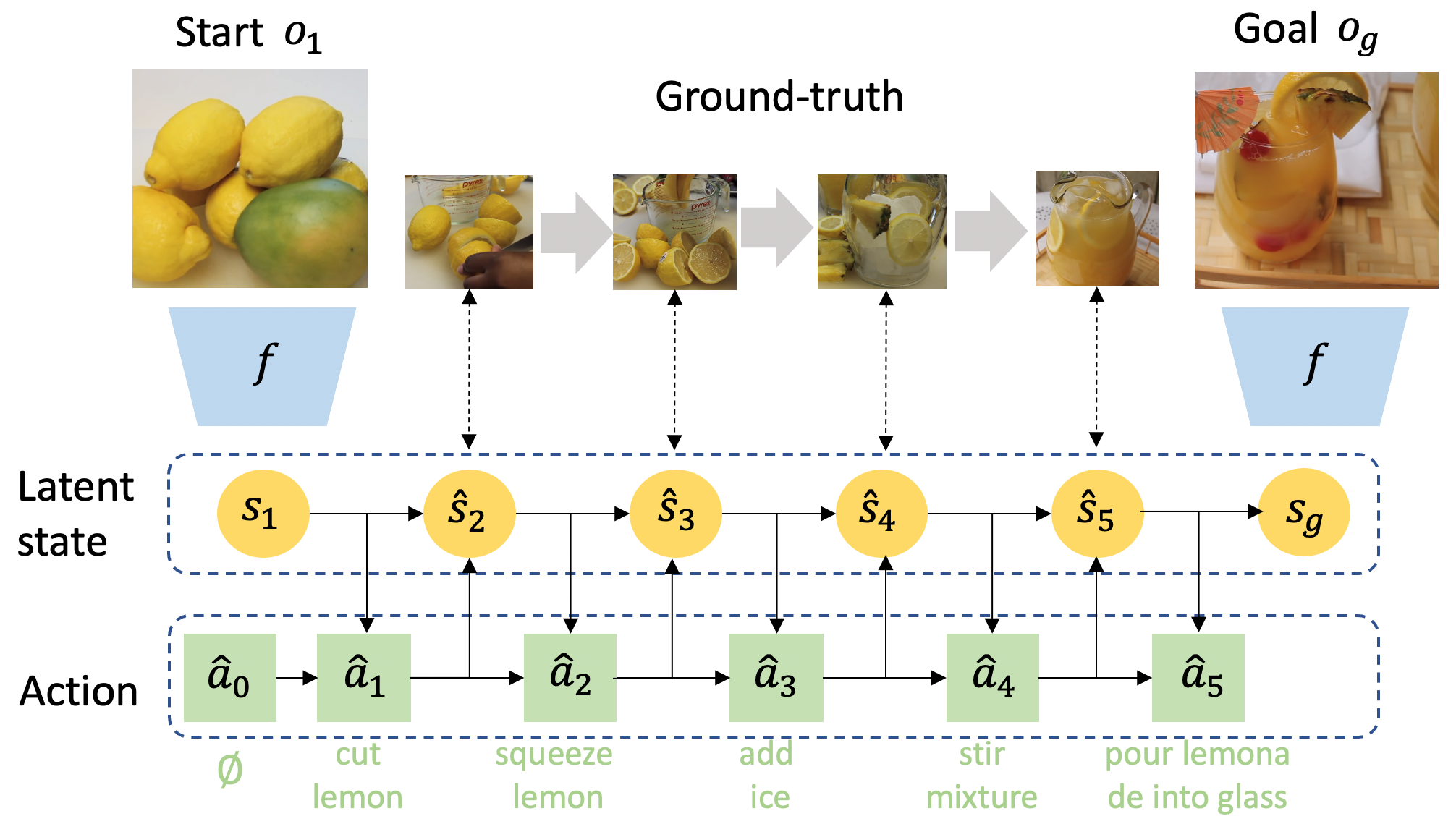}
  \caption{\textbf{\algoName Overview}. Given a visual observation as start and goal, the encoder $f(\cdot)$ extracts the feature about the planning trajectory. This transformer-based procedure planning model is responsible for learning plannable latent representations $\hat{s}$ and actions $\hat{a}$.}
\label{fig:overview}
\end{figure}

Procedure planning in instructional videos~\cite{chang2020procedure} (as shown in Figure~\ref{fig:overview}) aims to make goal-conditioned decisions by planning a sequence of high-level actions that can bring the agent from current observation to the goal.
Planning in instructional videos is a meaningful task since the ability to perform effective planning is crucial for an instruction-following agent.
Although learning from instructional videos is natural to humans, it is challenging for the AI system because it requires understanding human behaviors in the videos, focusing on actions and intentions. How to learn structured and plannable state and action spaces directly from unstructured videos is the key technical challenge of our task. 

It is crucial for autonomous agents to plan for complex tasks in everyday settings from visual observations~\cite{chang2020procedure}. Although reinforcement learning provides a powerful and general framework for decision making and control, its application in practice is often hindered by the need for extensive feature and reward engineering~\cite{9361118,Huang_DeepDecision_CoRL2020}. Moreover, deep RL algorithms are often sensitive to factors such as reward sparsity and magnitude, making well-performing reward functions particularly difficult to engineer. In many real-world applications, specifying a proper reward function is difficult. 

In this paper, we proposed a new framework for procedure planning from visual observations. We address these limitations with a new formulation of procedure planning and novel algorithms for modeling human behavior through a Transformer-based planning network~\algoName. Our method simultaneously learns the high-level action planning of assigned tasks and the representations of the decision-making process from human demonstrations.

We summarize our contributions as follows: 
\begin{itemize}
    \item We proposed a novel method, \underline{Pla}nning \underline{T}ransform\underline{e}r network (\algoName), for procedure planning in instructional videos task, which enjoys the advantage of long-term planning. We introduce some critical design choices that assist in learning cross-modal correspondence and, more importantly,  improve the accuracy of generated planning sequences. 
    \item We integrate Beam Search to \algoName to prevent it from large search discrepancies, and \jk{reduce} the performance degradation.
    \item Experimental results show that our framework outperforms the baselines in the procedure planning task on CrossTask, a real-world dataset. We also validated the possibility of applying procedural tasks on a real UR-5 platform.
\end{itemize}

\section{Related Work}

\noindent\textbf{Self-Attention and Transformer.}
Transformer-based architectures, eschew the use of recurrence in neural networks and instead trust entirely on self-attention mechanisms to draw global dependencies between inputs and outputs.
Self-attention~\cite{NIPS2017_3f5ee243} is particularly suitable for procedure planning, which can be seen as a sequence modeling task. Compared with Recurrent Neural Networks (RNNs), long short-term memory (LSTM)~\cite{10.1162/neco.1997.9.8.1735} and gated recurrent neural networks~\cite{chung2014empirical}, the advantages of self-attention includes avoiding compressing the whole past into a fixed-size hidden state, less total computational complexity per layer, and more parallelizable computations. In this paper, thanks to Transformers’ computational efficiency and scalability, we explore the possibility of marrying Transformer-based architecture for procedure planning.

\vspace{1mm}
\noindent\textbf{Learning to Plan from Pixels.}
Another related work is learning dynamics models for
model-based RL~\cite{hafner2019learning,fang2019dynamics,pmlr-v144-amos21a}. 
Recent works have shown that deep networks can learn to plan directly from pixel observations in domains such as table-top manipulation~\cite{NEURIPS2018_08aac6ac,srinivas2018universal,Chen_RED_ICRA20},
navigation~\cite{pathak2017curiosity,qiu2021egocentric,pan2020cross}, and locomotion in joint space~\cite{ehsani2018let}.
Universal Planning Networks (UPN)~\cite{srinivas2018universal} assumes the action space to be differentiable and uses a gradient descent planner to learn representations from expert demonstrations. 
\jk{Prior work~\cite{chang2020procedure} proposes the conjugate dynamics model to expedite the latent space learning, but suffers from compounding error. While they attempt to solve the same problem, however there exist some key methodical differences. Mainly, \algoName uses a transformer model to plan actions, which results in improved success rate, accuracy, and mIoU. 
Further, even with a good planner, due to the stochastic nature of predictions, planning is often low quality with a single roll-out. We address this issue with a beam search method.
The adversarial training scheme proposed by Bi et al.~\cite{bi2021procedure} is a complex, multi-stage pipeline. 
In contrast, we implement an end-to-end transformer architecture, simple but efficient.} 

\vspace{1mm}
\noindent\textbf{Learning from Instructional Videos.}
The interest has dramatically increased in recent years in understanding human behaviors by analyzing instructional videos~\cite{zhou2018towards,zhukov2019cross,9025668}. 
Event discovery tasks such as action recognition and temporal action segmentation~\cite{huang2016connectionist,blukis2022persistent,xu2018neural,huang2019neural},
state understanding~\cite{alayrac2017joint}, video prediction~\cite{yu2020modular} and video summarization / captioning~\cite{sun2019videobert,zhou2018towards} study recognition of human actions in video sequences. 
The other works~\cite{wang2019progressive} perform egocentric action anticipation model the relationships between past, future events, and incomplete observations.
Action label prediction~\cite{sener2019zero,Farha_2019_ICCV}
addresses the problem of anticipating all activities within a time horizon. However, the correct answer is often not unique due to the large uncertainty in human actions. 

\begin{figure*}[t]
\centering
  \includegraphics[width=0.63\linewidth]{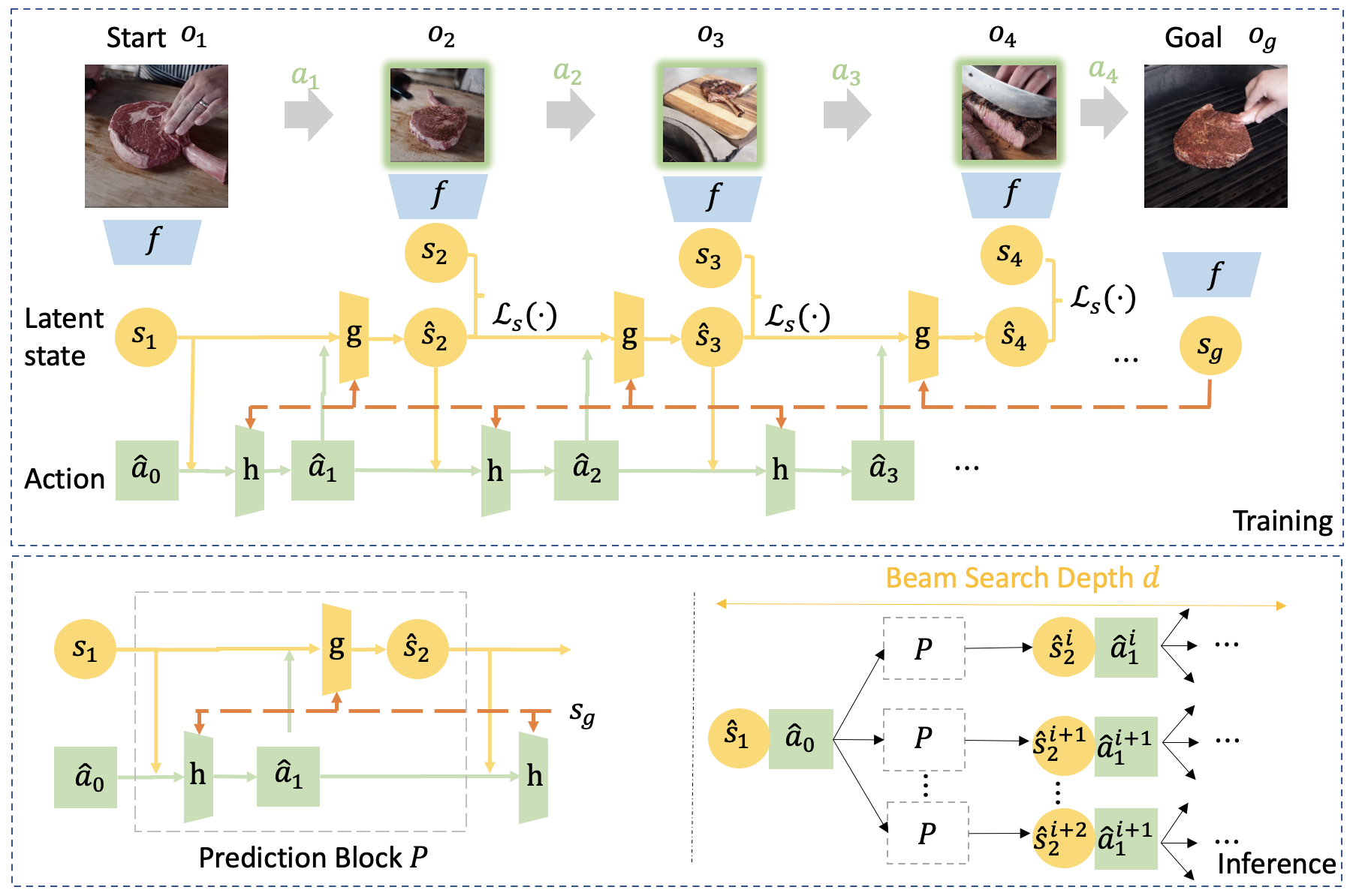}
  \includegraphics[width=0.30\linewidth]{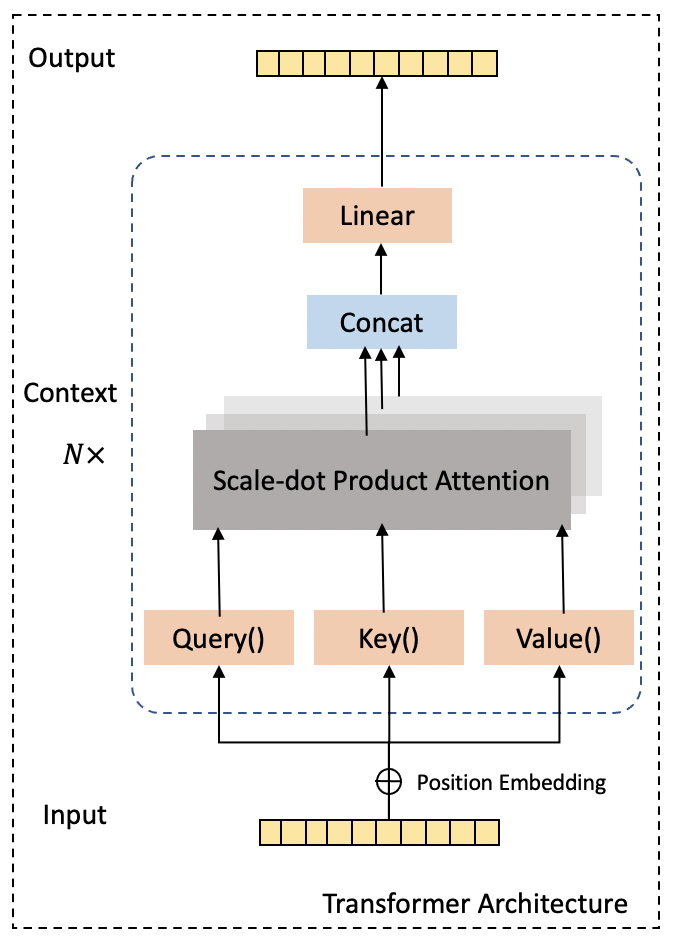}
  \caption{\textbf{\algoName Framework}. Given the start and goal visual observations, encoder $f$ outputs the latent representations $\hat{s}$. We set $\hat{a}_0=0$. The latent representation and predicted action are inferred using transformer-based action prediction model $h(\cdot)$ and state prediction model $g(\cdot)$. \jk{The goal state $s_g$ is an input to both models}. During training, the ground-truth action $a$ and state $s$ are given. During inference, we use Beam Search to enhance the trained $h(\cdot), g(\cdot)$. The right part shows the Transformer architecture we used for the planning model. \jk{Query(), Key(), and Value() are linear layers.}}
\label{fig:framework}
\end{figure*}

\section{Our Method: \algoName}
\subsection{Problem Setup}
We consider a similar setup to~\cite{chang2020procedure}: given the start visual observation $o_t$ and a visual goal $o_g$ that indicates the desired outcome for a particular task. During training, we have access to the observation-action pairs $\{(o_{t:g}, a_{t:g})\}\sim\pi_E$ that were collected by an expert attempting to reach the goals. When testing, only the start visual observation $o_t$ and a visual goal $o_g$ are given. Our objective is to plan a sequence of actions $[\hat{a}_t, \dots, \hat{a}_{t+T-1}]$ that can bring the underlying state of $o_t$ to that of $o_g$. $T$ is the horizon of planning, which means the number of task-level action steps the model is allowed to take. 
Figure~\ref{fig:overview}  shows a goal-oriented plannable example where the intermediate steps of performing a complex planning task are planned.

Our key insight is that the compounding error can be reduced by jointly learning the state and action representation with Transformer-based architecture.
As shown in the overall architecture in Figure~\ref{fig:framework}, 
the procedure planning problem $p(\hat{a}_{1:T}\vert o_1, \jk{o_g})$ is formulated as \jk{maximizing the probability}

\begin{footnotesize}
\begin{equation}
\begin{split}
    & p(\hat{a}_{1:T} \vert o_1, \jk{o_g}) \\
    = & \textrm{Pr}(\hat{s}_1 \vert o_1)\textrm{Pr}(\jk{\hat{s}_g} \vert \jk{o_g})\prod_{t=1}^{T} \textrm{Pr}(\hat{a}_t\vert \hat{s}_{t}, \hat{a}_{t-1}, \jk{\hat{s}_g}) \textrm{Pr}(\hat{s}_{t+1}\vert \hat{s}_{t}, \hat{a}_{t}, \jk{\hat{s}_g}).
    \label{equ:pp}
\end{split}
\end{equation}
\end{footnotesize}
In the following sections, we first discuss how to encode the latent semantic representation. Then, we will introduce how to solve the long-term procedure planning task with transformer-based architecture. Lastly, we will discuss how to apply learned representation to solve the procedure planning by integrating Beam Search.

\subsection{Latent Semantic Representation}
First, we use the state encoder $\hat{s} = f(o)$ that encodes the visual observation to a latent semantic representation, then added with learnable positional encoding, before they were input into the transformer layers.

The remaining question is: how to learn a planning model $p(\hat{a}_{1:T}, \hat{s}_{1:T} \vert \hat{s}_1, \jk{\hat{s}_g})$ to plan the action sequence and corresponding latent state representation? We assume the underlying process in Figure~\ref{fig:overview} is a fully observable goal-conditioned Markov Decision Process ($\mathcal{S, A, T}$), where $\mathcal{S}$ is the state space, $\mathcal{A}$ is the action space, $\mathcal{T:S\times A\rightarrow S}$ is the unknown transition probability distribution. 
We denote $h(\hat{a}_t \vert \hat{s}_t, \hat{a}_{t-1}, \jk{\hat{s}_g})$ as the \emph{action prediction model} conditioned on the current state, previous action, and goal, $g(\hat{s}_{t+1}\vert \hat{s}_{t}, \hat{a}_{t},  \jk{\hat{s}_g})$ as the \emph{state prediction model} conditioned on the previous state, goal, and previous action. They plan the sequence of actions and hidden states
from the initial state to the goal state. In this way, we are able to factorize the planning model $p(\hat{a}_{1:T}, \hat{s}_{1:T} \vert \hat{s}_1, \jk{\hat{s}_g})$ as:
\begin{equation}
\begin{split}
    & p(\hat{a}_{1:T}, \hat{s}_{1:T} \vert \hat{s}_1, \jk{\hat{s}_g}) \\
    = & \prod_{t=1}^T h(\hat{a}_t\vert \hat{s}_t, \hat{a}_{t-1}, \jk{\hat{s}_g})g(\hat{s}_{t+1}\vert \hat{s}_{t}, \hat{a}_{t}, \jk{\hat{s}_g}),
\end{split}
\end{equation}
where we use the convention that $\hat{s}_0 = 0$, $\hat{a}_0 = 0$. \jk{The data is in the form of tuples ($s_t$, $a_t$, $s_{t+1}$). We assume to start at $s_1$, and reach a goal $s_g$, in $T$ timesteps. We need to generate an action sequence $a_{1:T}$ of length $T$, which yields a state trajectory of length $T+1$.}

\subsection{Transition Transformer}
We propose a transformer-based network architecture that can learn the action-state correlation and generate planning sequences. The overview of this architecture is shown in Figure~\ref{fig:framework}. 
Our design choices are explained in detail below.

We introduce two cross-modal transformers: the action transformer $h(\hat{a}_t\vert \hat{s}_t, \hat{a}_{t-1},  \jk{\hat{s}_g})$, which learns the correspondence between previous action feature $\hat{a}_{t-1}$ and state feature $\hat{s}_t$, $\jk{\hat{s}_g}$ and generates the action prediction $\hat{a}_t$; the state transformer $g(\hat{s}_{t+1}\vert \hat{s}_{t}, \hat{a}_{t}, \jk{\hat{s}_g})$, which learns the correspondence between the state feature $\hat{s}_{t}$, $\jk{\hat{s}_g}$ and action feature $\hat{a}_{t}$ and generates the future state prediction $\hat{s}_{t+1}$. The model takes as input the whole sequence during training and all past pairs of state-action during inference, similarly to how transformer-like models usually work. 

Specifically, the output of the attention layer, the context vector $\bm{C}$ is computed using the query vector $\bm{Q}$ and the key $\bm{K}$ value $\bm{V}$ pair
from the input with an upper
triangular look-ahead mask $\bm{M}$ via

\begin{small}
\begin{equation}
\begin{split}
    \bm{C} & = {\rm Attn}(\bm{Q, K, V, M}) \\
    & = {\rm softmax} \left(\frac{\bm{QK}^T + \bm{M}}{\sqrt{D}}\right)\bm{V},
\end{split}
\end{equation}
\end{small}
where $D$ is the number of channels in the attention layer. The look-ahead-mask $\bm{M}$ is a triangular matrix \jk{to ensure that the predictions can depend only on the known outputs before}.

\subsection{Beam Search in Procedure Planning}
Given a transformer planning model \jk{$P_\theta$} parameterized by $\theta$ and an input $x$, which contains the information of current state, previous action step, and goal state, the problem of procedure planning task consists of finding a action sequence \jk{$\hat{\textbf{a}}$} such that \jk{$\hat{\textbf{a}} = \argmax_{\textbf{a}\in A}P_\theta(\textbf{a}\vert x)$}, where $A$ is the set of all sequences. \jk{$\textbf{a}=\{\hat{a}_1, \cdots, \hat{a}_{T}\}$} can be regarded as a sequence of tokens from vocabulary $\mathcal{V}$, where $T$ is the length of the sequence \jk{$\hat{\textbf{a}}$}. Then \jk{$P_\theta(\hat{\textbf{a}}\vert x)$} can be factored as
\begin{small}
\begin{equation}
\begin{split}
    \log P_\theta(\hat{\jk{\textbf{a}}}\vert x) & = \log \prod_{t=1}^{T}P_\theta(\hat{a}_t\vert \{\hat{s}_1, \hat{a}_0\};\cdots; \{\hat{s}_t, \hat{a}_{t-1}\}; \jk{\hat{s}_g}) \\
    & = \sum_{t=1}^{T}\log P_\theta(\hat{a}_t\vert \{\hat{s}_1, \hat{a}_0\}; \cdots; \{\hat{s}_t, \hat{a}_{t-1}\}; \jk{\hat{s}_g}).
\end{split}
\end{equation}
\end{small}

The discrepancy gap is the difference in log-probability between the most likely token and the chosen token~\cite{pmlr-v97-cohen19a,meister2020best}. At time step $t$, the discrepancy gap is
\begin{equation}
\begin{split}
    \max_{\hat{\jk{\textbf{a}}}\in\mathcal{V}} & [ \log P_\theta(\hat{\jk{\textbf{a}}}\vert \{\hat{s}_1, \hat{a}_0\};\cdots; \{\hat{s}_t, \hat{a}_{t-1}\}; \jk{\hat{s}_g}) \\
    & - \log P_\theta(\hat{a}_t\vert \{\hat{s}_1, \hat{a}_0\};\cdots, \{\hat{s}_t, \hat{a}_{t-1}\}; \jk{\hat{s}_g})].
\end{split}
\end{equation}
To avoid long-term procedure planning from significant search discrepancies, we introduce the discrepancy-constrained Beam Search during the \emph{inference phase of procedure planning}. 
The action log-probability output by action prediction model $h(\cdot)$ is used as the score function. The inference algorithm is shown as Algorithm~\ref{alg:beam_search}. In this way, we can \jk{reduce} the performance degradation. 

\subsection{Learning}
As shown in Figure~\ref{fig:framework}, we have three main components to optimize: state encoder $f$, action prediction model $h$, and state prediction model $g$.
We refer to the expert trajectory as $\tau^E = \{(s_t^E, a_t^E)\}$ and predicted trajectory $\tau=\{(\hat{s}_t, \hat{a}_t)\}$ as state-action pairs visited by the current planning model.

We optimize by descending the gradient in Equation~\ref{equ:state_action_learn}.
\begin{equation}
    \min_{\theta}\sum_{t=1}^T||\hat{s}_t - s_t^E||_2 + {\rm CE}(\hat{a}_t, a_t^E)
    \label{equ:state_action_learn}
\end{equation}
where ${\rm CE}$ is the cross-entropy loss.  In training, $T$-step sequence is output once.  In testing, single-step inference is made with Beam Search.
\begin{algorithm}[t] 
	\caption{\algoName: Planning Inference Phase}
	\label{alg:beam_search}
	\hspace*{\algorithmicindent} \textbf{Input}: sequence $\mathbf{x}$, \jk{Beginning Of Sequence $\rm{BOS}$, End Of Sequence $\rm{EOS}$}, buffer $B$, buffer size $\mathcal{N}$, score function $\textrm{score}(\cdot, \cdot)$, maximum sequence length $n_{\max}$, 
	maximum beam size $k$, planning model $p(\hat{s}_t^i, \hat{a}_t^i \vert \hat{s}_{t-1}^i, \hat{a}_{t-1}^i, \hat{s}_{T}^i)$
	\\
    \hspace*{\algorithmicindent} \textbf{Output}: searched sequence $B.\rm{max()}$
	\begin{algorithmic}
		\State $B_0 \leftarrow \{<0, \rm{BOS}>\}$
		\For{$t\in \{1, \cdots, n_{\max} - 1$\}}
		\State $B \leftarrow \emptyset$
		\For{$<w, (\hat{s}_{t-1}^i, \hat{a}_{t-1}^i, \hat{s}_{T}^i)> \in B_{t-1}$}
		\If {$\hat{s}_{t-1}^i.{\rm last()} = {\rm EOS}$}
		\State $B.{\rm add}(<w, (\hat{s}_{t-1}^i, \hat{a}_{t-1}^i, \hat{s}_{T}^i)>)$
		\State \textbf{continue}
		\EndIf
        \State $(\hat{s}_t^i, \hat{a}_t^i)\leftarrow p(\hat{s}_t^i, \hat{a}_t^i \vert \hat{s}_{t-1}^i, \hat{a}_{t-1}^i, \hat{s}_{T}^i)$
		\State $w \leftarrow {\rm score} (\hat{s}_t^i, \hat{a}_t^i, \hat{s}_{T}^i)$
		\State $B.{\rm add}(<w, (\hat{s}_t^i, \hat{a}_t^i, \hat{s}_{T}^i)>)$
		\EndFor
		\State $B_t\leftarrow B.{\rm top}(k)$
		\EndFor \\
		\Return{$B.\rm{max()}$}
	\end{algorithmic}
\end{algorithm}

\section{Experiments}
In our experiments, we aim to answer the following questions: (1) Is \algoName efficient and scalable to procedure planning tasks? (2) Can \algoName learn to plan on the interactive environment? 
To answer Question 1, we evaluate \algoName on CrossTask, a real-world offline instructional video dataset. We show procedure planning with our algorithm performs better on the CrossTask dataset than previous methods. To answer Question 2, we evaluate our method on a real-world UR-5 robot arm platform. 

\begin{figure*}[t]
\centering
  \includegraphics[width=0.9\linewidth]{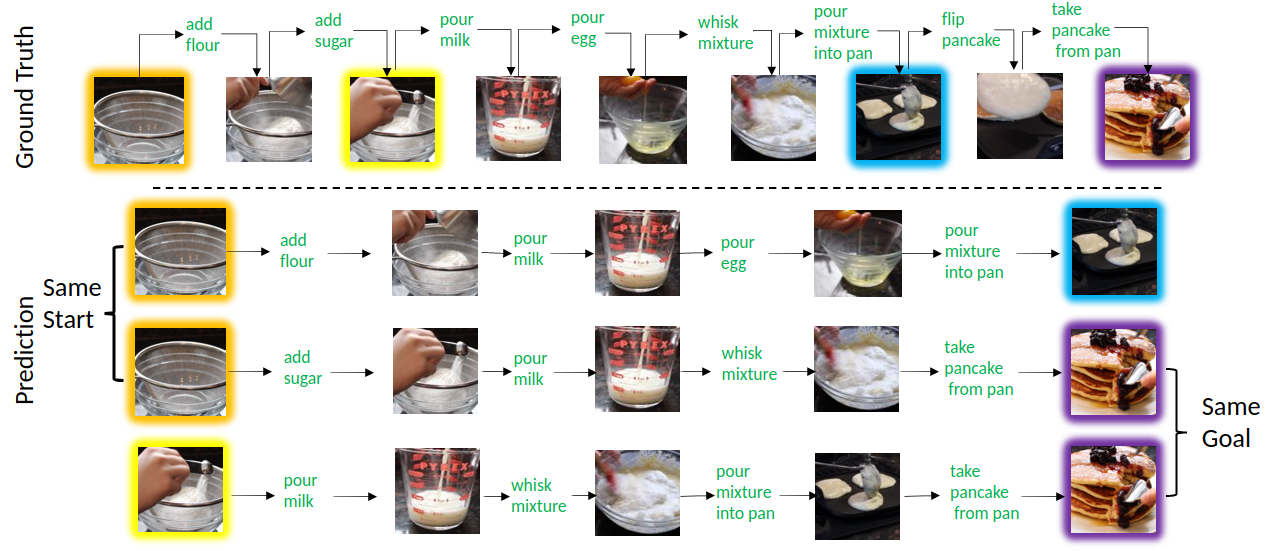}
  \caption{\textbf{Qualitative results of procedure planning on CrossTask}. Qualitative results of procedure planning for \emph{Make Pancakes}. The top row describes the correct action sequence required to "make pancakes". To examine our method's robustness, We vary the start and goal observations to evaluate our method. The results show that our approach is robust to perform planning within different stages in the video.}
\label{fig:pp_qualitative_crosstask}
\vspace{-5pt}
\end{figure*}

\begin{table*}
\begin{center}
\caption{CrossTask Results. Our model outperforms baselines in terms of success rate, accuracy, and mIoU.}
\label{tab:crosstask}
\begin{tabular}{l|c|c|c|c|c|c}
\toprule
\multirow{2}{*}{Method} & \multicolumn{3}{c|}{Prediction Length $T=3$} & \multicolumn{3}{c}{Prediction Length $T=4$} \\
\cline{2-4}\cline{5-7}
& Success Rate (\%) & Accuracy (\%) & mIoU (\%) & Success Rate (\%) & Accuracy (\%) & mIoU (\%) \\
\midrule
Random & $<$0.01  & 0.94  & 1.66  &  $<$0.01  &  0.83  & 1.66  \\
RB~\cite{sun2019videobert} & 8.05 & 23.30 & 32.06 & 3.95 &  22.22 & 36.97 \\
WLTDO~\cite{ehsani2018let} & 1.87 & 21.64 & 31.70 & 0.77 & 17.92 & 26.43 \\
UAAA~\cite{Farha_2019_ICCV} & 2.15 & 20.21 & 30.87 & 0.98 & 19.86 & 27.09 \\
UPN~\cite{srinivas2018universal} & 2.89 & 24.39 & 31.56 & 1.19 & 21.59 & 27.85 \\
DDN~\cite{chang2020procedure} & 12.18 & 31.29 & 47.48 & 5.97 & 27.10 & 48.46 \\
\midrule
\algoName (Ours) & $\bm{16.00}$ & $\bm{36.17}$ & $\bm{65.91}$ & $\bm{14.00}$ & $\bm{35.29}$ & $\bm{55.36}$ \\
\bottomrule
\end{tabular}
\end{center}
\end{table*}

\subsection{Experimental Setup}
\noindent\textbf{Datasets.}
We first evaluate \algoName on an instructional video dataset CrossTask~\cite{zhukov2019cross}. For real-world UR-5 experiments, we collect a \emph{UR-5 Reaching Dataset} which consists of 100 trajectories (2150 first-person-view RGB image and corresponding action pairs) as a training set and evaluate on a real UR-5 platform. 

\noindent\textbf{Implementation Details.} We compare with the baselines in~\cite{chang2020procedure} with the metrics such as success rate, accuracy, and mIoU. 
We use the Transformer architecture~\cite{NIPS2017_3f5ee243} as the transition model with $8$ self-attention layers and $8$ heads. The transition model is two-headed: one for action prediction 
the other for state prediction.
\jk{Let $L_s$ be the hidden dimension for state and $L_a$ be the embedding dimension for action.}
The state encoder in our model is two fully-connected layers with [64, $L_s$] units in each layer and Leaky-ReLU as non-linearity function. 
\jk{We encode $(o_t, o_g)$ to be $L_s$-dim $s_t$. $a_t$ is encoded (in transformer terminology: "tokenized") to be a $L_a$-dim embedding and concatenated together with $s_t$ to be provided as input to the transformer. The size of the input of the transformer is $L_s + L_a$ and the size of the output of the transformer is also $L_s + L_a$. $s_{t+1}$ is the first $L_s$-dim features split from transformer output. $a_{t+1}$ is decoded from the rest of transformer output.  In our experiments, $L_s=32$ and $L_a$ is the total number of possible actions. For CrossTask experiments, $L_a=105$.} One-hot vectors are used for this action classification purpose.
During training, all models are optimized by Adam~\cite{kingma2014adam} with the starting learning rate of $10^{-4}$. We train our model for 200 epochs on a single GTX 1080 Ti GPU. 

\subsection{Evaluating Procedure Planning on CrossTask}
First, we choose a real-world instructional video dataset CrossTask~\cite{zhukov2019cross} to conduct our experiments. CrossTask comprises $2,750$ videos ($212$ hours). Each video depicts one of the $18$ primary long-horizon tasks.
To test the trained agent’s generalization capability, for the videos in each task, we randomly divide the videos in each task into $70\%/30\%$ splits for training and testing. 
Each video can be regarded as a sequence of images $V_i$ (where $i$ is the index of frames) that have annotated with a sequence of action labels $a_j$ and each action starts at frame index $s_j$ and ends at frame index $e_j$.
Same as the setup of~\cite{chang2020procedure}: we choose frames around the beginning of the captions $V_{s_t-\delta/2:s_t +\delta/2}$ as $o_t$ , caption description $a_t$ as the semantic meaning of action, and images nearby the end $V_{e_t-\delta/2:e_t +\delta/2}$ as the next observation $o_{t+1}$. Here, $\delta$ controls the duration of each observation, and we set $\delta = 2$ for all data we have used in our paper.
Our state-space $\mathcal{S}$ is the pre-computed features provided in CrossTask: each second of the video is encoded into a $3,200$-dimensional feature vector
, which is a concatenation of the I3D~\cite{8099985}, Resnet-152~\cite{he2016deep}, and audio VGG features~\cite{7952132}. 
The action space $\mathcal{A}$ is constructed by enumerating all combinations of predicates and objects, which provides $105$ action labels and is shared across all $18$ tasks. 
Our method is suitable for modeling longer trajectories, but we restrict the experiments to horizontal lengths $T=3\sim 4$ to maintain a consistent comparison with state-of-the-art methods.

Recall that in procedure planning, given the start and goal observations $o_1$ and \jk{$o_g$}, the agent needs to output a valid procedure $\{a_1, \cdots, a_T\}$ to reach the specified goal. 
As illustrated in Table~\ref{tab:crosstask}, as instructional videos’ action space is not continuous, the gradient-based planner of UPN cannot work well. By introducing Beam Search, our \algoName has a better performance in terms of success rate, accuracy, and mIoU. By designing a model with transformer-based components, we show that our model outperforms all the baseline approaches on real-world videos.

In Figure~\ref{fig:pp_qualitative_crosstask}, we visualize some examples of the predicted procedure planning results on CrossTask, where the task is to \emph{Make Pancake}. Our model is able to predict a sequence of actions with correct ordering. Specifically, the most challenging step in \emph{Make Pancake} is the "add flour" and "add sugar" step, where visual differences are not significant, and it can only be inferred from context and sequence relationships. 

\begin{table}[t]
\begin{center}
\caption{Success Rate (\%) of UR5.}
\label{tab:actionet}
\begin{tabular}{l|c|c}
\toprule
Method &  \multicolumn{2}{c}{UR5} \\
\midrule
Prediction Length &  $T=3$ &  $T=4$ \\
\midrule
Random  & $<$0.01    & $<$0.01   \\
RB~\cite{sun2019videobert}  & 32 & 26 \\
RL~\cite{chang2020procedure} & 50 & 44  \\
WLTDO~\cite{ehsani2018let}  & 42 & 38 \\
UAAA~\cite{Farha_2019_ICCV}  & 44 & 40 \\
UPN~\cite{srinivas2018universal}  & 44 & 38 \\
DDN~\cite{chang2020procedure} & 52 & 46  \\
\midrule
\algoName (Ours) &$\bm{60}$  & $\bm{52}$  \\
\bottomrule
\end{tabular}
\end{center}
\end{table}

\begin{figure*}[t]
\centering
  \includegraphics[width=0.29\linewidth]{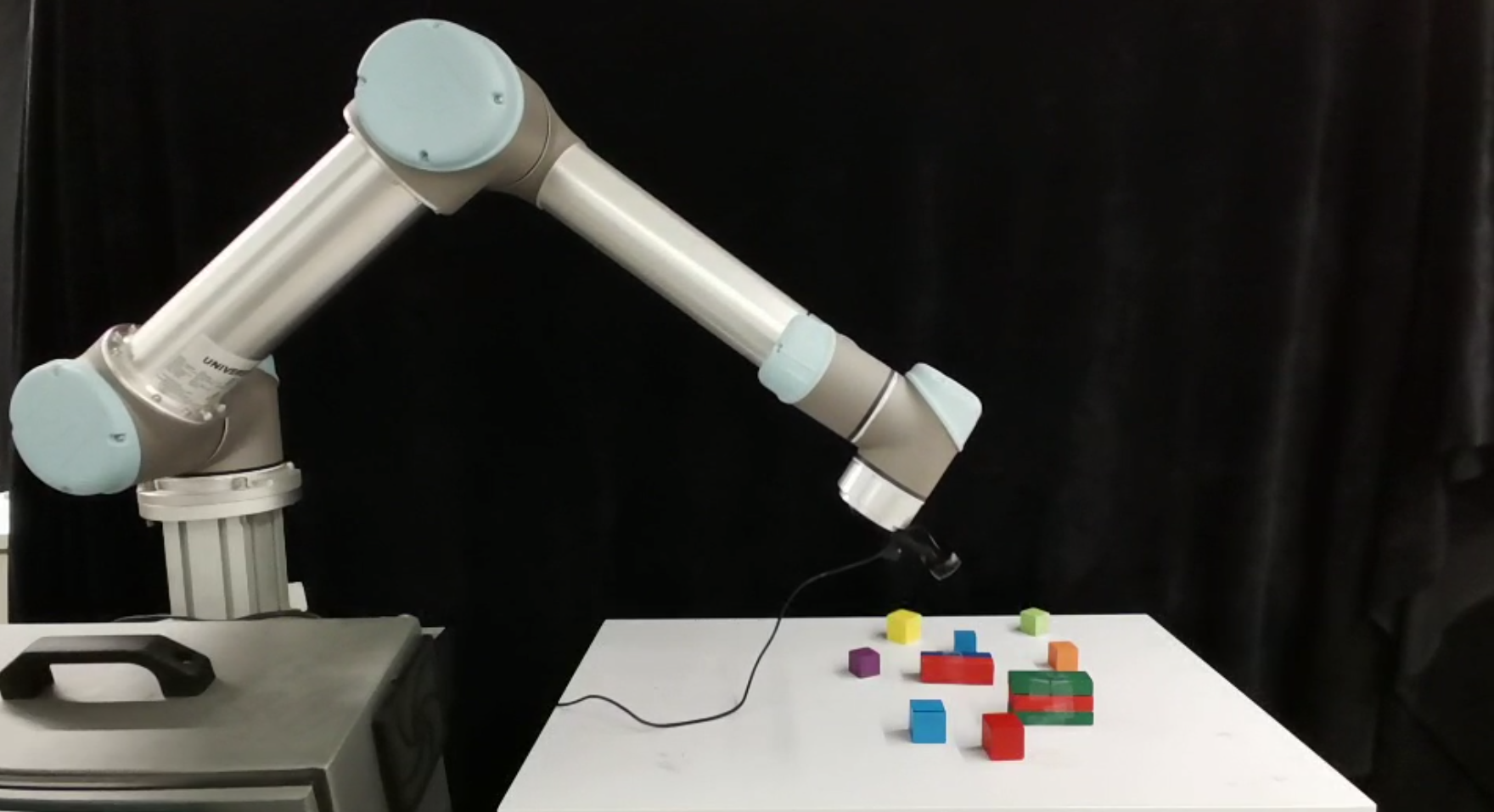}
  \includegraphics[width=0.6\linewidth]{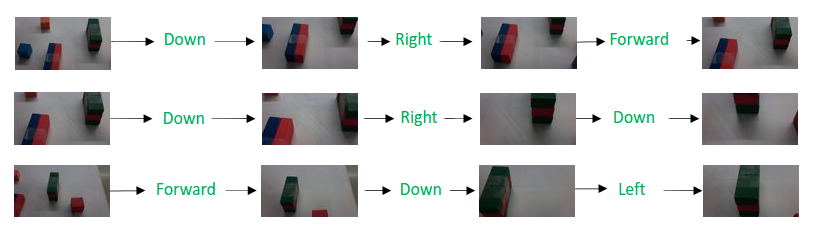}
  \caption{\textbf{Qualitative results of procedure planning on UR5}.}
\label{fig:pp_ur5}
\end{figure*}

\subsection{Evaluating Procedure Planning on Real Robot}
Previous Procedure Planning research has rarely reported experimental results in real robots. There remains a gap between offline training and real-world applications. To validate the possibility of applying procedural tasks in the real environment, we conduct experiments on "Reaching a block" (cf. Figure \ref{fig:pp_ur5}) using a Universal Robot UR5 system. \jk{While this task is easy in simulation, it can be difficult for real robot~\cite{gu2017deep}}. 
Our agent learns to achieve tasks by imitating expert demonstrations. After the RGB-D camera is installed in front of the manipulator, the observation space includes the raw RGB images at the current position and goal position. The action space includes \texttt{None, Up, Down, Left, Right, Forward, Backward}. UR5 Reacher consists of episodes of interactions, where each episode is $T$ time steps long. The fingertip of UR5 Reacher is confined within a 3-dimensional 0.7m × 0.5m × 0.4m boundary. The robot is also constrained within a joint-angular boundary to avoid self-collision.

In the task of Reaching, the robot is required to start at the current observation and then move to a goal observation. Since the controller of the robot is imperfect, we consider a reach to be successful if the robot reaches within $5$cm of the block.
The UR5 robotic arm is controlled by human volunteers to reach the target, and thus 100 offline expert demonstration trajectories are generated. All methods are evaluated on a real UR5 robotic arm for 50 episodes. As shown in Table~\ref{tab:actionet}, our approach outperforms the other methods. \jk{Even though other baselines have succeeded in the offline dataset}, real robotic reaching lags far behind human performance and remains unsolved in the field of robot learning.

\section{Conclusion} 
To conclude, we propose a cross-modal transformer-based architecture to address the procedure planning problem, which can capture long-term time dependencies. Moreover, We propose to enhance the transformer-based planner with Beam Search. Finally, we evaluate our method on a real-world instructional video dataset. The results indicate that our method can learn a meaningful action sequence for planning and recover the human decision-making process. We also validated the possibility of applying procedural tasks on a real UR-5 platform.

\section*{Acknowledgment}
Animesh Garg is supported in part by CIFAR AI Chair at the Vector Institute for AI and NSERC Discovery Award.

\ifCLASSOPTIONcaptionsoff
  \newpage
\fi



%

{\small
\bibliographystyle{IEEEtran}
\bibliography{IEEEexample}
}

%




\end{document}